\documentclass[acmtog,nonacm]{acmart}

\usepackage{booktabs} % For formal tables

\usepackage[ruled]{algorithm2e} % For algorithms

\SetAlFnt{\small}
\SetAlCapFnt{\small}
\SetAlCapNameFnt{\small}
\SetAlCapHSkip{0pt}

\setcopyright{none}
\renewcommand\footnotetextcopyrightpermission[1]{}

\begin{document}
% Title portion
\title{RealSkin: Spatio-Spectral Partial Neural Adjoint Maps for Image-to-3D Attribute Transfer}

% DO NOT ENTER AUTHOR INFORMATION FOR ANONYMOUS TECHNICAL PAPER SUBMISSIONS TO SIGGRAPH 2019!
\author{Jing Li}
\orcid{0009-0008-9292-9832}
\affiliation{%
  \institution{Jilin University}
%  \streetaddress{104 Jamestown Rd}
%  \city{Williamsburg}
%  \state{VA}
%  \postcode{23185}
  \country{China}}
\email{lij25@mails.jlu.edu.cn}
\author{Yawei Luo}
\authornote{Corresponding authors.}
\affiliation{%
  \institution{Zhejiang University}
  \country{China}
}
% \email{yaweiluo@zju.edu.cn}
\author{Xiangze Meng}
\affiliation{%
 \institution{Jilin University}
 \country{China}}
%\email{aprna_patel@rguhs.ac.in}
\author{Ying Li}
\affiliation{%
 \institution{North China University of Technology}
%  \streetaddress{30 Shuangqing Rd}
 \country{China}
}
%\email{chan0345@tsinghua.edu.cn}
\author{Tieru Wu}
\affiliation{%
 \institution{Jilin University}
%  \city{Prague}
 \country{China}}
%\email{yanting02@gmail.com}
\author{Rui Ma}
\authornotemark[1]
\affiliation{%
 \institution{Jilin University}
%  \department{School of Engineering}
%  \city{Charlottesville}
%  \state{VA}
%  \postcode{22903}
 \country{China}
}
%\affiliation{%
%  \institution{University of Minnesota}
%\email{tinghe@uva.edu}
%\author{Chengdu Huang}
%\author{John A. Stankovic}
%\author{Tarek F. Abdelzaher}
%\affiliation{%
%  \institution{University of Virginia}
%  \department{School of Engineering}
%  \city{Charlottesville}
%  \state{VA}
%  \postcode{22903}
%  \country{USA}
%}

\renewcommand\shortauthors{Li, J. et al}

\makeatletter
\let\@authorsaddresses\@empty
\makeatother

\begin{abstract}
Creating photorealistic 3D assets requires bridging the appearance gap between real-world observations and synthetic models. A promising approach is to transfer visual attributes from real images onto synthetic 3D surfaces. Traditional methods struggle with resolution mismatch and the inherent discreteness of point correspondences. In contrast, resolution-robust functional maps enable smooth attribute propagation but rely on near-isometry assumptions and topological consistency. To address these limitations, we propose RealSkin, a self-supervised framework that performs correspondence optimization in a learned spectral domain, guided by spatial correspondences. We first introduce a spatial-guided registration algorithm to establish coarse correspondences under severe topological discrepancies. To relax strict isometric assumptions and handle partial correspondences, we further design a spectral-aware neural adjoint network that incorporates partial correspondences into a neural function space and models non-isometric residuals for correspondence refinement. Experimental results demonstrate that our method achieves state-of-the-art performance on challenging real-to-synthetic scenarios. The code will be publicly released.
\end{abstract}





\begin{teaserfigure}
  \includegraphics[width=\textwidth]{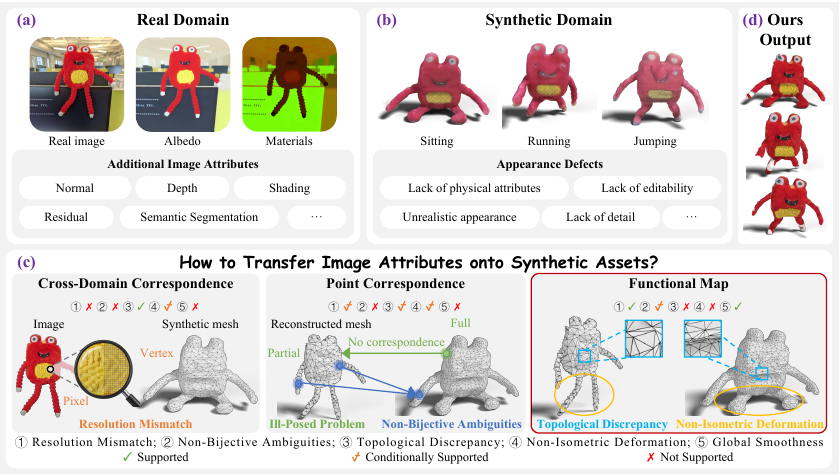}
  \centering
  \caption{RealSkin addresses topological discrepancies and non-isometric deformations within a functional map framework, enabling high-fidelity transfer of appearance attributes from images to 3D assets. (a)–(b) illustrate the substantial appearance gap between real and synthetic data. (c) illustrates the advantages, limitations, and challenges of different appearance transfer methods, which motivates us to choose and improve a functional map-based approach. (d) shows our high-fidelity transfer results.}
  % \Description{Enjoying the baseball game from the third-base
  % seats. Ichiro Suzuki preparing to bat.}
  \label{fig:teaser}
\end{teaserfigure}

\maketitle

\section{Introduction}

Physically realistic modeling constitutes a fundamental and long-standing problem in computer vision and graphics, with applications ranging from digital content production to world models \cite{team2025hunyuanworld} and spatial computing \cite{jun2025dr}. In the context of 3D assets, this objective centers on three core dimensions: geometry \cite{li2025craftsman3d}, appearance \cite{yu2024texgen}, and physical dynamics \cite{chen2025physgen3d,zhan2025physiopt,cao2025physx}. Among these, appearance serves as the critical determinant of visual believability. Although recent generative models \cite{xiang2025structured,li2025triposg,xiang2026native} have significantly improved geometric quality and basic texture synthesis, producing locally photorealistic appearance that faithfully reflects real-world imagery remains a challenge.

Unlike generative approaches, we aim to bridge the appearance gap between real-world observations and synthetic 3D assets through direct image attribute transfer. This requires fine-grained alignment of visual attributes across domains with fundamentally different sampling patterns and structural representations. Directly establishing correspondences between image pixels and 3D geometry is often hindered by resolution mismatch. Alternatively, lifting 2D observations into 3D and constructing point-wise correspondences is inherently discrete, making smooth attribute transfer difficult.

In this work, we revisit appearance transfer from a shape correspondence perspective. Our key insight is that, instead of establishing point-wise correspondences, we reconstruct local geometry from the reference image and seek a functional correspondence \cite{ovsjanikov2012functional} between the two shapes. However, directly applying standard functional maps in real-to-synthetic appearance transfer is challenging for two fundamental reasons. First, standard formulations rely on isometric assumptions and struggle in the presence of \textbf{\textit{topological discrepancies}} and \textbf{\textit{non-isometric deformations}}, as shown in Fig.~\ref{fig:teaser}. Second, the reference image provides only \textbf{\textit{partial observations}}. Establishing correspondences from the full synthetic geometry to the observed reference is intrinsically ill-posed.

To address these limitations, we restrict the correspondence domain to regions where valid correspondences are likely to exist and extract a partial sub-region of the synthetic asset via ray casting. We then formulate appearance transfer as a partial functional map between the reconstructed reference surface and this sub-mesh.

Based on this formulation, we propose RealSkin, a novel self-supervised optimization framework for real-to-synthetic appearance transfer. It consists of two key components. First, we introduce \textbf{spatial-guided registration} (SGR) algorithm to compute a coarse correspondence under topological discrepancies. We cast point matching as a constrained assignment problem over hybrid geometric descriptors with progressive spatial expansion. This step provides a stable initialization under significant structural mismatch. Second, we design a \textbf{spectral-aware neural adjoint network} (SANAN), which projects partial correspondences into a neural functional space for refinement. It jointly models isometric structure and non-isometric residuals, relaxing the strict isometric assumption in classical functional maps. This provides a stable optimization space for refining dense correspondences under non-isometric deformations. Extensive experiments demonstrate that RealSkin successfully imparts a photorealistic appearance to the local surfaces of synthetic assets, effectively bridging the real-to-synthetic appearance gap.

In summary, our key contributions are as follows:
% itemize
\begin{itemize}
\item We formulate real-to-synthetic appearance transfer as a partial functional correspondence problem between image-reconstructed surfaces and synthetic geometry.
\item We present RealSkin, a novel self-supervised framework for reliable surface attribute transfer from real-world images to synthetic 3D assets.
\item We propose SGR algorithm, a robust coarse correspondence method that handles severe topological discrepancies via spatially constrained assignment.
\item We design SANAN, which embeds partial correspondences into a spectral functional space and models both isometric structure and non-isometric residuals for stable refinement.
\end{itemize}

% Head 1
\section{Related Works}
% Head 2
\subsection{Appearance Enhancement}
% \paragraph{Appearance Enhancement.}
Enhancing the visual realism of synthetic 3D assets remains a pivotal challenge in 3D content creation. Meta 3D TextureGen \cite{bensadoun2024meta} achieves sharper textures via UV-space inpainting, but is constrained by the scarcity of high-quality UV data. Several recent methods \cite{li2024dreamtexture, luo20253denhancer, hunyuan3d2025hunyuan3d} leverage 2D diffusion models to improve multi-view image quality for texture enhancement. However, these methods perform diffusion in a latent space, which often leads to loss of fine-grained details and requires strict image–geometry alignment. Unlike these works, our method directly transfers attributes from images, enabling better preservation of fine-grained details from the reference image.

\begin{figure*}
  \centering
  \includegraphics[width=0.9\linewidth]{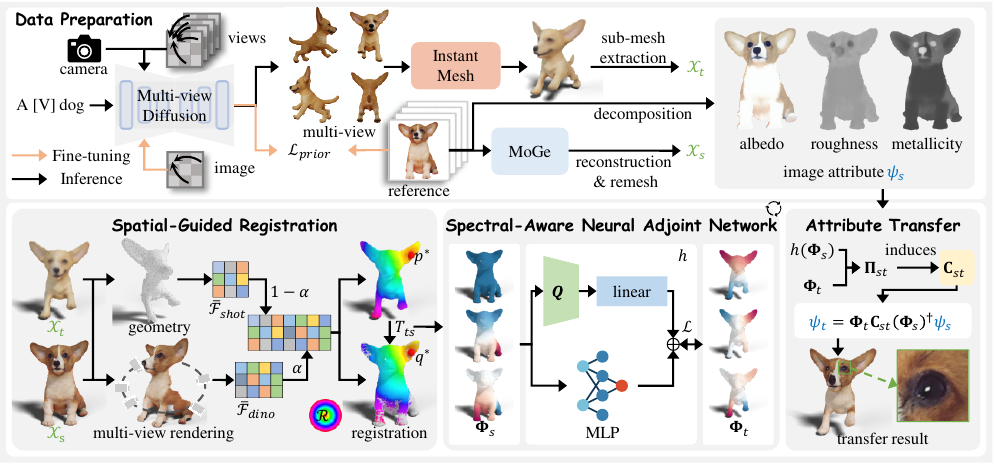}
  \caption{Overview of the RealSkin framework. Given a reference image, we reconstruct the source surface and synthesize a corresponding 3D asset, from which a target sub-mesh is extracted. Coarse correspondences are first established using SGR, and then refined in a neural functional space via SANAN to obtain accurate functional maps. These maps enable smooth signal transfer between the reference and target domains. Image attributes are flexibly defined depending on the task setting.}
  \label{fig:pipeline}
\end{figure*}

\subsection{Shape Correspondence and Registration}
% \paragraph{Shape Correspondence and Registration.}
Estimating correspondences between 3D shapes is a fundamental problem in geometry processing. Recent approaches generally fall into two main paradigms: registration methods and functional map methods. Registration methods typically deform the source to resemble the target shape, and the correspondences are retrieved simply by nearest-neighbor search in the embedding space \cite{ezuz2019elastic, sundararaman2024deformation}. Although recent works \cite{roetzer2024spidermatch, liu2025stable} seek to jointly optimize deformation and matching with guaranteed optimality, their scalability remains constrained.

Alternatively, the functional map framework \cite{ovsjanikov2012functional} formulates shape correspondence as a linear mapping between function spaces. Building upon this, traditional methods like PFM \cite{rodola2017partial} and FSPSM \cite{litany2017fully} address partiality in the spectral space by exploiting the insight that the eigenbases of partial shapes manifest within those of full ones. However, these methods are fundamentally constrained by strict isometric assumptions. While numerous learning-based approaches \cite{litany2017deep, cao2023unsupervised, donati2020deep, sharp2022diffusionnet, li2024deformable, attaiki2021dpfm, xie2025echomatch} have been proposed, their reliance on training data severely limits their generalization to in-the-wild scenarios. Recently, NAM \cite{vigano2025nam} circumvented this data dependency through neural representations, yet it remains confined to full-shape matching tasks. In contrast, our approach combines spatial guidance with neural spectral optimization, achieving robust performance in partial correspondence scenarios under severe topological variations.

\subsection{2D Models for 3D Tasks}
% \paragraph{2D Models for 3D Tasks.}
Recently, 2D vision foundation models \cite{simeoni2025dinov3,radford2021learning, kirillov2023segment, rombach2022high} have achieved remarkable progress, driven by the availability of large-scale datasets. In contrast, the development of 3D vision foundation models remains hindered by severe data scarcity. Therefore, recent approaches \cite{fu2024geowizard, shi2023zero123++} have increasingly leveraged the strong generalization capabilities of 2D models to tackle 3D tasks. Marigold \cite{ke2025marigold} fine-tunes diffusion models for dense image analysis. SRIF \cite{sun2024srif} employs DiffMorpher \cite{zhang2024diffmorpher} to distill 2D image interpolation into 3D morphing, and DIFF3F \cite{dutt2024diffusion} aggregates 2D features onto 3D surfaces to enhance robustness. Inspired by these advancements, we incorporate 2D features to enhance spatial correspondences, providing a reliable initialization for our subsequent optimization in the neural spectral space.

\section{Preliminaries}

% \subsection{Laplace-Beltrami Operator}
\paragraph{Laplace-Beltrami Operator.}
For functional correspondence, we base our consistent spectral representation of geometric signals on the Laplace-Beltrami operator (LBO). Let $\mathcal{X}$ be a compact 2-manifold. The intrinsic gradient $\nabla_{\mathcal{X}}$ and the positive semi-definite LBO $\Delta_{\mathcal{X}}$ generalize the corresponding notions from flat spaces to manifolds. The Laplacian admits an eigendecomposition $\Delta_{\mathcal{X}} \phi_i(x) = \lambda_i \phi_i(x)$, for $x\in \mathcal{X}$. Here, $0 = \lambda_1 \leq \lambda_2 \leq \dots$ are the eigenvalues and $\{\phi_i\}_{i \geq 1}$ are the corresponding eigenfunctions. The eigenfunctions form an orthonormal basis of $L^2(\mathcal{X})$ generalizing the classical Fourier analysis, allowing any function $f \in L^2(\mathcal{X})$ to be expanded into the Fourier series as $f(x) = \sum_{i \geq 1} \langle f, \phi_i \rangle_{\mathcal{X}} \phi_i(x)$.

% \subsection{Functional Correspondence}
\paragraph{Functional Correspondence.}
Given two manifolds $\mathcal{X}_1$ and $\mathcal{X}_2$, a correspondence $T_{12}: \mathcal{X}_1 \to \mathcal{X}_2$ induces via pull-back a linear map \cite{ovsjanikov2012functional} $T_{21}^F: L^2(\mathcal{X}_2) \to L^2(\mathcal{X}_1)$. Let $\mathbf{\Phi}_1$ and $\mathbf{\Phi}_2$ be orthonormal bases of $L^2(\mathcal{X}_1)$ and $L^2(\mathcal{X}_2)$, respectively. The functional map $T_{21}^F$ can be encoded as a matrix $\mathbf{C}_{21}$ that maps the spectral coefficients. In the discrete setting, given the matrix representation $\mathbf{\Pi}_{21}$ of $T_{12}$, the functional map can be computed as
\begin{equation}
    \mathbf{C}_{21} = \mathbf{\Phi}_1^\dagger \mathbf{\Pi}_{21} \mathbf{\Phi}_2,
    \label{eq: fm}
\end{equation}
where $\dagger$ denotes the Moore–Penrose pseudoinverse. The functional map $T_{21}^F$ induces the adjoint operator \cite{huang2017adjoint} $T_{12}^A: L^2(\mathcal{X}_1) \to L^2(\mathcal{X}_2)$, which propagates functions in the forward direction and satisfies $\langle T_{12}^A f, g \rangle_{L^2(\mathcal{X}_2)} = \langle f, T_{21}^F g \rangle_{L^2(\mathcal{X}_1)}$ for any $f \in L^2(\mathcal{X}_1)$ and $g \in L^2(\mathcal{X}_2)$. In the case of orthonormal bases, as with LBO, $\mathbf{A}_{12} = \mathbf{C}_{21}^\top$.

To recover the point correspondence $T_{12}$, points are modeled as Dirac delta distributions, whose spectral coefficients correspond to the rows of $\mathbf{\Phi}$. The correspondence is extracted via nearest neighbor (NN) search between the spectral embeddings
\begin{equation}
    \label{eq: nn_fm}
    T_{12} = \text{NN}(\mathbf{\Phi}_1, \mathbf{\Phi}_2 \mathbf{C}_{21}^\top).
\end{equation}

Considering LBO truncated basis $\mathbf{\Phi}_1^k$ and $\mathbf{\Phi}_2^k$, Eq.~\ref{eq: nn_fm} can be reformulated using the adjoint operator as $T_{12} = \text{NN}(\mathbf{\Phi}_1^k, \mathbf{\Phi}_2^k \mathbf{A}_{12})$, which is optimized via the following energy functional \cite{melzi2019zoomout}:
\begin{equation}
    \mathcal{E}(\mathbf{\Pi}_{21}, \mathbf{A}_{12}) = \| \mathbf{\Pi}_{21} \mathbf{\Phi}_2^k \mathbf{A}_{12} - \mathbf{\Phi}_1^k \|_F^2,
    \label{eq: adjoint operator}
\end{equation}
where $\| \cdot \|_F$ indicates the matrix Frobenius-norm. In our method, functional correspondence forms the core computational backbone for RealSkin.

\section{Methodology}

Fig.~\ref{fig:pipeline} provides an overview of the RealSkin framework. We begin in Sec.~\ref{sec: preparation} by formalizing the attribute transfer problem and describing the data preparation process. Sec.\ref{sec: registration} then introduces our spatial-guided progressive registration algorithm, which establishes reliable initial point-wise correspondences based on extrinsic features and spatial cues. Finally, Sec.\ref{sec:spectral_nam} presents our spectral-aware neural adjoint network for accurate attribute transfer.

\subsection{Problem Formulation and Data Preparation}
\label{sec: preparation}

% \subsubsection{Problem Formulation}
\paragraph{Problem Formulation.}
Given a real-world image $\mathbf{I} \in \mathbb{R}^{H \times W \times 3}$ and a synthetic 3D asset $\mathcal{X}_\text{syn}$ depicting the same object, our primary objective is to transfer the appearance attributes of $\mathbf{I}$ onto the surface of $\mathcal{X}_\text{syn}$. 

We formulate this task within the functional map framework. Specifically, we reconstruct the local geometry from $\mathbf{I}$ to serve as the source mesh $\mathcal{X}_s$. The attributes can be defined as a function $\psi_s: \mathcal{X}_s \to \mathbb{R}^c$, where $c$ is the dimension of the signal. Due to the partial nature of $\mathcal{X}_s$, establishing a mapping from $\mathcal{X}_\text{syn}$ to this partial mesh is highly ill-posed. To make this problem tractable, we extract a target sub-mesh $\mathcal{X}_t \subset \mathcal{X}_\text{syn}$ via ray casting. Conceptually, we treat $\mathcal{X}_s$ as a complete domain and $\mathcal{X}_t$ as a local region. This effectively transforms the problem into a tractable partial-to-full mapping, formally denoted as $T: \mathcal{X}_t \to \mathcal{X}_s$.

Given the mapping $T$, we can transport the attribute function $\psi_s$ defined on $\mathcal{X}_s$ to a function $\psi_t$ defined on $\mathcal{X}_t$ via the composition $\psi_t = \psi_s \circ T$. Specifically, this implies:
\begin{equation}
  \psi_t(p) = \psi_s(T(p)), \quad \forall p \in \mathcal{X}_t.
  \label{eq: transport}
\end{equation}

This formulation introduces several challenges, including partiality, topological discrepancies, and non-isometric deformation between $\mathcal{X}_s$ and $\mathcal{X}_t$.

% \subsubsection{Data Preparation}
\paragraph{Data Preparation.}
To reconstruct $\mathcal{X}_s$, we first utilize a foreground segmentation model \cite{zheng2024bilateral} to extract a precise mask from the reference image $\mathbf{I}$. A partial surface is then recovered over the masked region via MoGe \cite{wang2025moge}. Additionally, we consider intrinsic image decomposition~\cite{ke2025marigold} as a way to conceptually define physical attributes on $\mathcal{X}_s$. For appearance transfer, we directly adopt the reference image as the texture signal.

To synthesize $\mathcal{X}_\text{syn}$, we first fine-tune a text-conditioned multi-view diffusion model \cite{shi2024mvdream} to enable subject-driven multi-view image generation. Our objective function consists of two terms: (a) an image diffusion term for subject-prior embedding and (b) a parameter preservation term for maintaining multi-view ability, which is computed as:
\begin{equation}
  \mathcal{L}_\text{prior}(\theta,\mathbf{I}_\text{id})=\mathbb{E}_{\epsilon,t} ||\epsilon - \epsilon_{\theta}(\mathbf{I}_\text{id}, t)||^2_2 + \mu \frac{\|\theta - \theta_0\|_1}{N_\theta},
  \label{eq: prior}
\end{equation}
where $\mathbf{I}_\text{id}$ denotes the subject images, $t$ is the diffusion time step, $\epsilon$ is the added noise, $\theta_0$ indicates the initial parameters of original multi-view diffusion, $N_\theta$ is the number of parameters, and $\mu$ is a balancing parameter set to 1. During this fine-tuning process, we replace the cross-view self-attention with intra-image self-attention, effectively assimilating identity priors from a collection of 2D reference images. At inference, given a textual prompt, we restore the cross-view self-attention mechanism to generate subject-driven multi-view images. These outputs are fed into InstantMesh \cite{xu2024instantmesh} to model the 3D asset. Next, we extract the target sub-mesh $\mathcal{X}_t$ via ray casting and perform a coarse-scale geometric calibration on $\mathcal{X}_s$. Additional implementation details are provided in the supplementary material.

\subsection{Spatial-Guided Registration}
\label{sec: registration}

% \subsubsection{Hybrid Extrinsic Descriptors}
\paragraph{Hybrid Extrinsic Descriptors.}

We begin by establishing dense correspondences between the target mesh $\mathcal{X}_t$ and the source mesh $\mathcal{X}_s$, consisting of $N_t$ and $N_s$ vertices respectively. Considering the unreliability of extrinsic geometric descriptors under severe topological discrepancies, we incorporate semantic priors to enhance correspondence robustness. Specifically, we compute the SHOT descriptor \cite{tombari2010unique}, denoted as $\mathcal{F}_\text{shot}$, on the mesh surface to capture local 3D geometric cues. To construct the semantic component, we render the mesh from multiple viewpoints and extract DINO \cite{simeoni2025dinov3} features. Following \cite{dutt2024diffusion}, we leverage the known camera parameters to unproject these image-space features back onto the 3D vertices, yielding semantic features $\mathcal{F}_v$ from the $v$-th viewpoint. For each vertex observed from a set of $V$ valid views, we aggregate these features via average pooling to derive the semantic descriptor $\mathcal{F}_\text{dino} = \frac{1}{V} \sum_{v=1}^{V} \mathcal{F}_v$.

We then employ a feature fusion strategy \cite{zhang2023tale} that concatenates the $L_2$-normalized geometric ($\bar{\mathcal{F}}_\text{shot}$) and semantic ($\bar{\mathcal{F}}_\text{dino}$) features:
\begin{equation}
    \mathcal{F}_\text{hybrid} := [\alpha \bar{\mathcal{F}}_\text{dino}; (1-\alpha)\bar{\mathcal{F}}_\text{shot}],
\end{equation}
where $\alpha$ is a balancing weight set to $0.5$ in all experiments.

% \subsubsection{Geodesic Ring Decomposition}
\paragraph{Geodesic Ring Decomposition.}
\begin{figure}
    \centering
    \includegraphics[width=0.9\linewidth]{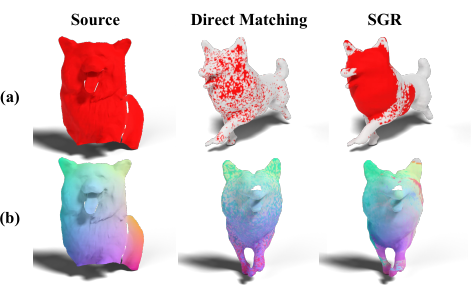}
    \caption{Effectiveness of SGR. (a) Direct matching based on maximal similarity suffers from severe spatial incoherence. SGR effectively mitigates this issue, while simultaneously providing a reliable reference for sub-mesh extraction. (b) Comparison of color transfer results. Compared to direct matching, SGR enables smoother and more accurate transfer from the source mesh.}
    \label{fig: SGR}
\end{figure}

Given the hybrid descriptors, we first compute a cross-shape similarity matrix $\mathbf{G} \in \mathbb{R}^{N_t \times N_s}$ between $\mathcal{X}_t$ and $\mathcal{X}_s$. Since direct maximum-similarity matching often yields severe spatial incoherence, as shown in Fig.~\ref{fig: SGR}, we instead introduce a spatial-guided matching strategy. We first identify an optimal seed pair $(p^* \in \mathcal{X}_t, q^* \in \mathcal{X}_s)$. Starting from these anchors, we compute the geodesic distance field on $\mathcal{X}_t$, denoted as $d_t(p)$. Subsequently, we partition the surface of $\mathcal{X}_t$ into a sequence of concentric geodesic rings based on uniform distance intervals. The $j$-th geodesic ring $\mathcal{R}_j^t$, for $j \in \{0, 1, \dots, J-1\}$, is defined as
\begin{equation}
\mathcal{R}_j^t := \{ p \in \mathcal{X}_t \mid j\omega \leq d_t(p) < (j+1) \omega \}, \text{where } \omega = \frac{\max_{p \in \mathcal{X}_t} d_{t}(p)}{J},
\label{eq: target ring}
\end{equation}
and $J$ is the total number of rings. Using the same interval $\omega$, we apply an identical partitioning scheme to $\mathcal{X}_s$ around the source anchor $q^*$ to obtain the source rings $\mathcal{R}_j^s$. More details are provided in the supplementary material.

\begin{figure}
    \centering
    \includegraphics[width=\linewidth]{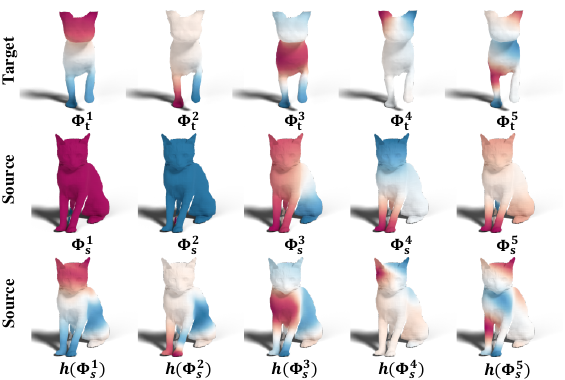}
    \caption{The top and middle rows display the first five eigenfunctions on the target and source shapes, respectively; note the inconsistent behavior at corresponding indices. The bottom row illustrates the basis functions optimized by SANAN: the new basis manifests the same behavior as in the first row, and is at the same time localized on the latent corresponding part.}
    \label{fig: Sanan}
\end{figure}

% \subsubsection{Progressive Ring-Constrained Registration}
\paragraph{Spatially Constrained Progressive Registration.}

With the geodesic ring decomposition, we formulate dense matching as a progressive assignment problem. At the $j$-th stage, instead of searching over all vertices in $\mathcal{X}_s$, we construct a dynamically relaxed candidate pool $\mathcal{D}_j^s$. This pool consists of vertices in the current source ring $\mathcal{R}_j^s$ and unmatched vertices from the preceding $\tau$ rings. For a target vertex $p \in \mathcal{R}_j^t$ and a source candidate $q \in \mathcal{D}_j^s$, we define a cost matrix $\mathbf{E}$ that evaluates both feature similarity and spatial consistency:
\begin{equation}
    \mathbf{E}_{p, q} = -\left( \beta_1 \mathbf{G}_{p, q} + \beta_2 \left\langle \hat{\mathbf{d}}^t_{p}, \hat{\mathbf{d}}^s_{q} \right\rangle + \beta_3 \exp\left( -\frac{\left| \rho_t(p) - \rho_s(q)\right|}{\rho_t(p) + \rho_s(q) + \iota} \right) \right),
    \label{eq: score}
\end{equation}
where $\mathbf{G}_{p, q}$ is the initial feature similarity between $p$ and $q$; $\hat{\mathbf{d}}^t_{p}$ and $\hat{\mathbf{d}}^s_{q}$ denote the normalized relative directional vectors from their respective anchors; $\rho_t(p)$ and $\rho_s(q)$ represent the radial Euclidean distances from the anchors $p^*$ and $q^*$, respectively; $\iota$ is a small constant to ensure numerical stability; and the empirical weights $\beta_1 = 0.3, \beta_2 = 0.4, \beta_3 = 0.3$ balance the three metric components.

Finally, we perform the linear sum assignment over $\mathbf{E}$ via the Hungarian algorithm. Iterating this process across all rings yields a dense point-wise mapping, represented by the correspondence matrix $\mathbf{\Pi}_{st} \in \{0, 1\}^{N_t \times N_s}$.

\subsection{Spectral-Aware Neural Adjoint Network}
\label{sec:spectral_nam}

Although the SGR establishes a reliable coarse mapping $\mathbf{\Pi}_{st}$, the inherent non-bijective ambiguities and vertex discreteness render point-wise correspondences ill-suited for direct attribute transfer. Therefore, we further refine the correspondences within the functional map framework.

% \subsubsection{Theoretical Formulation}
\paragraph{Theoretical Formulation.}
To this end, we initially consider the partial correspondence task by assuming a near-isometric correspondence between a latent sub-region of $\mathcal{X}_s$ and $\mathcal{X}_t$. However, the global support of Laplacian eigenfunctions makes them highly unstable in the presence of missing regions, resulting in substantial basis misalignment. Inspired by \cite{litany2017fully}, instead of explicitly masking the spatial domain, we absorb the partiality into a basis transformation $\mathbf{Q} \in \mathbb{R}^{k \times r}$ to construct a set of quasi-harmonic bases $\hat{\mathbf{\Phi}}_s^r = \mathbf{\Phi}_s^k\mathbf{Q}$. This ortho-projection allows the transformed basis to naturally approximate standard Laplacian eigenfunctions within the corresponding region. With this design, the adjoint matrix is adapted to the reduced localized space as $\mathbf{A}_{ts} \in \mathbb{R}^{r \times r}$, and Eq.~\ref{eq: adjoint operator} can thus be reformulated as:
\begin{equation}
    \mathcal{E}(\mathbf{\Pi}_{st}, \mathbf{A}_{ts}, \mathbf{Q}) = \|\mathbf{\Pi}_{st}\mathbf{\Phi}_s^k\mathbf{Q}\mathbf{A}_{ts} - \mathbf{\Phi}_t^r\|_F^2,
    \label{eq: projection}
\end{equation}
where $r$ is dynamically determined by the number of eigenvalues on $\mathcal{X}_t$ that are strictly smaller than the maximum truncated eigenvalue of $\mathcal{X}_s$, and $\mathbf{Q}$ resides on the Stiefel manifold. This formulation ensures the optimization remains fully spectral, bypassing the need for vertex-dependent spatial indicator functions.
% where $\mathbf{Q}$ resides on the Stiefel manifold, effectively compressing the high-dimensional spectral representation of the source into a localized subspace. This formulation ensures the optimization remains fully spectral, bypassing the need for vertex-dependent spatial indicator functions.

Interpreting the spectral bases as feature embeddings, the above discussions imply that we can consider $\mathbf{Q}\mathbf{A}_{ts}$ as a regularized way of aligning the embeddings to minimize Eq.~\ref{eq: projection}. However, since linear spectral mappings often struggle with complex topological discrepancies and significant non-isometric deformations, we extend this formulation by leveraging neural networks to parameterize a broader correspondence space.

% \subsubsection{Neural Representation}
\paragraph{Neural Representation.}

To generalize the action of the orthogonal projection and the adjoint operator to functional spaces approximated by truncated embeddings, we define $\mathcal{H}$ as the following space of neural functions:
\begin{equation}
    \mathcal{H} = \left\{ h: \mathbb{R}^k \to \mathbb{R}^r \ \middle| \ h(\mathbf{y}) = (\mathbf{y}\mathbf{Q})\mathbf{A} + \sigma(\sigma(\mathbf{y}\mathbf{W}^{(1)})\mathbf{W}^{(2)})\mathbf{W}^{(3)} \right\},
    \label{eq: network}
\end{equation}
where $\mathbf{Q} \in \mathbb{R}^{k \times r}$ is an orthogonal projection layer, $\sigma$ is an activation function, $\mathbf{A} \in \mathbb{R}^{r \times r}$ is a learnable linear transformation explicitly designed to parameterize the theoretical adjoint matrix $\mathbf{A}_{ts}$, $\mathbf{W}^{(1)} \in \mathbb{R}^{k \times K}$, $\mathbf{W}^{(2)} \in \mathbb{R}^{K \times K}$, and $\mathbf{W}^{(3)} \in \mathbb{R}^{K \times r}$ are linear parametric transformations, with $k \neq K $.

Given the truncated spectral bases $\mathbf{\Phi}_s^k \in \mathbb{R}^{N_s \times k}$ and $\mathbf{\Phi}_t^r \in \mathbb{R}^{N_t \times r}$, where $r$ is estimated from $k$, along with the source eigenvalue diagonal matrix $\mathbf{\Lambda}_s^k \in \mathbb{R}^{k \times k}$, we can interpret $\mathcal{H}$ as a space of functions between these spectral bases. Therefore, leveraging the previously established correspondence matrix $\mathbf{\Pi}_{st}$, we can optimize the neural function $h \in \mathcal{H}$ by minimizing the alignment energy:
\begin{equation}
    \mathcal{L}_\text{align} = \| h(\mathbf{\Pi}_{st}\mathbf{\Phi}_s^k) - \mathbf{\Phi}_t^r \|_F^2.
    \label{eq: align}
\end{equation}

As expressed in Eq.~\ref{eq: network}, the linear branch extracts a spectral subspace to model isometric mappings, while the non-linear residual absorbs non-isometric deviations. Furthermore, to preserve intrinsic shape smoothness, we define a quasi-harmonic loss as:
\begin{equation}
    \mathcal{L}_{\text{qh}} = \text{off}( \mathbf{Q}^\top \mathbf{\Lambda}_s^k \mathbf{Q}),
    \label{eq: qh_reg}
\end{equation}
where $\text{off}(\cdot)$ denotes the sum of squared off-diagonal elements. This penalty encourages the learned basis to approximate the Laplacian eigenfunctions.

We further impose a cross-frequency decoupling regularization on the linear operator $\mathbf{A}$:
\begin{equation}
   \mathcal{L}_\text{cf} = \text{off}(\mathbf{A}).
   \label{eq: cf_reg}
\end{equation}
Together with Eq.~\ref{eq: qh_reg}, this establishes a structured factorization where $\mathbf{Q}$ serves as a spectral projector and $\mathbf{A}$ models the adjoint operator of the isometry, driving the non-linear residual to focus on complex, non-isometric mappings.

The final loss is formulated as follows:
\begin{equation}
    \mathcal{L} = \gamma_1 \mathcal{L}_{\text{align}} + \gamma_2 \mathcal{L}_{\text{qh}} + \gamma_3 \mathcal{L}_{\text{cf}} ,
    \label{eq: loss}
\end{equation}
where $\gamma_1$, $\gamma_2$, and $\gamma_3$ are scalar hyperparameters weighting the respective structural priors.

\begin{figure}
    \centering
    \includegraphics[width=1.0\linewidth]{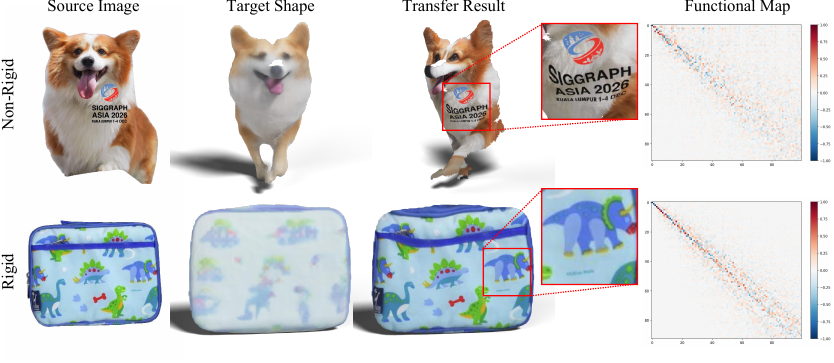}
    \caption{Visualization of texture transfer on both rigid and non-rigid objects. Our method produces high-quality results. The resulting functional maps exhibit a slanted diagonal structure, indicating that the learned correspondences preserve intrinsic geometric consistency.}
    \label{fig: result}
\end{figure}

\begin{algorithm}[t]
\SetAlgoNoLine
\KwIn{$\mathbf{\Pi}_{st}$}
\KwOut{$\mathbf{\Pi}_{st}$} 

\For{$k_{ini} \leq k \leq k_{end}$}{
    Optimize for $h$ minimizing Eq.~\ref{eq: loss} \;
    Compute $\mathbf{\Pi}_{st} = \text{NN}(\mathbf{\Phi}_t^r, h(\mathbf{\Phi}_s^k))$ \;
}
\caption{Iterative Refinement}
\label{alg:one}
\end{algorithm}
% \subsubsection{Iterative Refinement and Attribute Transfer}
\paragraph{Iterative Refinement and Attribute Transfer.}
We adopt a Neural ZoomOut (NZO) spectral refinement strategy following \cite{vigano2025nam}, as detailed in Algorithm~\ref{alg:one}. The refinement is performed in a self-supervised manner, progressively improving spectral alignment without any ground-truth correspondence annotations. Leveraging the refined correspondence matrix $\mathbf{\Pi}_{st}$, we compute the functional map $\mathbf{C}_{st}$ via least-squares optimization as in Eq.~\ref{eq: fm}. Subsequently, a signal $\psi_s$ defined on $\mathcal{X}_s$ can be transferred to $\mathcal{X}_t$ as $\psi_t$ by mapping its spectral coefficients through $\mathbf{C}_{st}$:
\begin{equation}
    \psi_t = \mathbf{\Phi}_t^r \mathbf{C}_{st} (\mathbf{\Phi}_s^k)^\dagger \psi_s.
    \label{eq: transfer}
\end{equation}

\section{Experiments}

\begin{table}
  \centering
  \caption{Quantitative comparison. We evaluate our method against shape correspondence methods and generative appearance enhancement methods. Our method yields superior performance across both visual and geometric metrics. $^*$ indicates metrics aligned with MaterialMVP \cite{he2025materialmvp}. $^\ddagger$ denotes generative appearance enhancement methods, while others are functional map-based. Best results in \textbf{bold}, second best \underline{underlined}.}
  \label{tab:combined_comparisons}
  \resizebox{\columnwidth}{!}{
  % \begin{tabular*}{\textwidth}{@{\extracolsep{\fill}} l ccccc ccc @{}}
  \begin{tabular}{l ccccc ccc}
    \toprule
    & \multicolumn{5}{c}{\textbf{Appearance}} & \multicolumn{3}{c}{\textbf{Geometry}} \\
    \cmidrule(lr){2-6} \cmidrule(lr){7-9}
    \textbf{Method} & SSIM $\uparrow$ & PSNR $\uparrow$ & LPIPS $\downarrow$ & CLIP $\uparrow$ & DINO $\uparrow$ & mCD $\downarrow$ & pCD $\downarrow$ & Flip\% $\downarrow$ \\
    \midrule
    PFM \cite{rodola2017partial} & 0.8362 & 15.81 & 0.1543 & 0.7238 & 0.2083 & 5.485 & 5.665 $\pm$ 0.86 & 11.75 $\pm$ 2.17 \\
    DPFM \cite{attaiki2021dpfm} & 0.8505 & 15.14 & 0.1527 & 0.7376 & 0.1747 & 8.576 & 10.059 $\pm$ 4.51 & \underline{8.62 $\pm$ 4.35} \\
    EchoMatch \cite{xie2025echomatch} & 0.8430 & 17.18 & 0.1454 & 0.7927 & 0.3878 & 6.270 & 6.615 $\pm$ 1.29 & 31.66 $\pm$ 2.98 \\
    DIFF3F \cite{dutt2024diffusion} & \underline{0.8579} & 17.04 & 0.1475 & 0.8103 & 0.3370 & 5.956 & 5.896 $\pm$ 1.17 & 9.55 $\pm$ 3.71 \\
    NAM \cite{vigano2025nam} & 0.8539 & \underline{18.29} & \underline{0.1305} & \underline{0.9269} & \underline{0.7246} & \underline{2.733} & \underline{2.784 $\pm$ 0.13} & 12.99 $\pm$ 11.25 \\

    \midrule
    UniTEX$^\ddagger$ \cite{liang2025unitex} & 0.8481 & 17.66 & 0.1321 & \underline{0.9112} & \underline{0.7823} & --- & --- & --- \\
    Hunyuan3D 2.1$^{*,\ddagger}$ \cite{hunyuan3d2025hunyuan3d} & 0.8493 & 17.68 & \underline{0.1302} & 0.9071 & 0.7698 & --- & --- & --- \\
    LumiTex$^\ddagger$ \cite{bao2025lumitex} & \underline{0.8547} & \underline{17.81} & 0.1303 & 0.8990 & 0.7706 & --- & --- & --- \\
    \midrule
    \textbf{Ours} & \textbf{0.8581} & \textbf{18.32} & \textbf{0.1263} & \textbf{0.9416} & \textbf{0.8527} & \textbf{2.632} & \textbf{2.647 $\pm$ 0.10} & \textbf{4.04 $\pm$ 1.91} \\
    \bottomrule
    \end{tabular}}
  % \end{tabular*}
\end{table}

\subsection{Implementation Details}
% \paragraph{Implementation Details.}
We fine-tune the multi-view diffusion model~\cite{shi2024mvdream} on subject-specific images with white backgrounds. In this stage, we revert to 2D attention and train the model for 600 steps with a batch size of 4, utilizing a learning rate of $2 \times 10^{-6}$. For SANAN, we set $\gamma_1 = 1$, $\gamma_2 = 3 \times 10^{-2}$, and $\gamma_3 = 1$. During the upsampling stage, we set $k_{\text{ini}} = 10$ and $k_{\text{end}} = 100$ with a step size of 18. The nonlinear module consists of two layers with a hidden dimension of 128.

\subsection{Evaluation Datasets and Metrics}
% \paragraph{Evaluation Datasets and Metrics.}
We randomly construct 40 shape pairs from the Google Scanned Objects \cite{downs2022google} and DreamBooth \cite{ruiz2023dreambooth} datasets for evaluation. The selected pairs exhibit substantial geometric diversity, spanning non-rigid animals and rigid objects with fundamentally different structural properties. We directly use the reference image as appearance attribute for transfer experiments. To evaluate the accuracy of the functional map $\mathbf{C}_{st}$, we compare the rendered target sub-meshes against the reference images using standard visual metrics: SSIM, PSNR, LPIPS, CLIP~\cite{radford2021learning}, and DINO~\cite{simeoni2025dinov3}. Geometrically, we assess the smoothness of the point-wise correspondences $\mathbf{\Pi}_{st}$ via Conformal Distortion (CD)~\cite{ezuz2019reversible}. For each target face, CD is defined as $\kappa = \varsigma_1/\varsigma_2+\varsigma_2/\varsigma_1$, where $\varsigma_i$ are the singular values of the Jacobian of the mapping to its image under $\mathbf{\Pi}_{st}$. We report three geometric statistics: (i) mCD, the global median of per-face conformal distortion $\kappa$ pooled over all target triangles across pairs; (ii) pCD, the per-pair median of $\kappa$, reported as mean ± std across pairs; and (iii) Flip, the per-pair area fraction of reversed-orientation triangles, summarized across pairs as mean ± std.

\subsection{Comparison Methods}
% \paragraph{Comparison Methods.}
To evaluate the attribute transfer performance of RealSkin, we compare our method against two primary categories of baselines. First, we consider shape correspondence methods, including: (1) optimization-based PFM \cite{rodola2017partial}; (2) learning-based functional map methods, DPFM \cite{attaiki2021dpfm} and EchoMatch \cite{xie2025echomatch}; (3) the zero-shot descriptor-based DIFF3F \cite{dutt2024diffusion}; and (4) the test-time optimization approach NAM \cite{vigano2025nam}. Second, we benchmark against state-of-the-art generative models for appearance enhancement, including UniTEX \cite{liang2025unitex} for shaded texture synthesis, as well as Hunyuan3D 2.1 \cite{hunyuan3d2025hunyuan3d} and LumiTex \cite{bao2025lumitex} for high-fidelity PBR material generation. Note that Hunyuan3D 2.1 is enhanced with MaterialMVP \cite{he2025materialmvp}. All experiments are conducted on a single NVIDIA RTX 4090 GPU.

\subsection{Comparative Studies}
% \paragraph{Comparative Studies.}
Compared with existing shape correspondence methods, our approach achieves superior performance across both visual and geometric metrics (Tab.~\ref{tab:combined_comparisons}). From a geometric perspective, our method significantly reduces triangle inversions (Flip\%) and achieves lower conformal distortion ($\kappa$). As shown in Fig.~\ref{fig: Comparison_conformal}, the distortion distribution is more concentrated and closer to the ideal value of $\kappa = 2$, indicating improved angle preservation and mapping stability. Qualitatively, our method exhibits less stretching and distortion, producing more plausible texture transfer results (Fig.~\ref{fig: qualitative_comparison}). We further present qualitative comparisons of physical attribute transfer in Fig.~\ref{fig: attribute_transfer} and demonstrate a downstream relighting application enabled by the transferred attributes in Fig.~\ref{fig: relighting}.

When evaluated against recent generative appearance enhancement methods, our approach also achieves superior quantitative performance (Tab.~\ref{tab:combined_comparisons}). Unlike generative baselines, which synthesize appearance implicitly, our method explicitly transfers image attributes, enabling more faithful preservation of fine-grained details from the reference image (Fig.~\ref{fig: qualitative_comparison_gen}).

\begin{table}
  \caption{Quantitative ablation study. Best results are highlighted in bold.}
  \label{ablation-wo-table}
  \centering
  \resizebox{\columnwidth}{!}{
  \begin{tabular}{l ccccc ccc}
    \toprule
    & \multicolumn{5}{c} {\textbf{Appearance}} & \multicolumn{3}{c}{\textbf{Geometry}} \\
    \cmidrule(rl){2-6} \cmidrule{7-9}
    \textbf{Model} & SSIM $\uparrow$ & PSNR $\uparrow$ & LPIPS $\downarrow$ & CLIP $\uparrow$ & DINO $\uparrow$
    & mCD $\downarrow$ & pCD $\downarrow$ & Flip\% $\downarrow$ \\
    \midrule
    RealSkin (Full) & \textbf{0.8581} & \textbf{18.32} & \textbf{0.1263} & \textbf{0.9416} & \textbf{0.8527} & \textbf{2.632} & \textbf{2.647 $\pm$ 0.10} & \textbf{4.04 $\pm$ 1.91}\\
    \midrule
    w/o DINO & 0.8465 & 17.62 & 0.1389 & 0.8907 & 0.5242 & 2.903 & 2.899 $\pm$ 0.12 & 9.25 $\pm$ 4.07 \\
    w/o SGR & 0.8562 & 18.27 & 0.1299 & 0.9216 & 0.7250 & 2.840 & 2.857 $\pm$ 0.12 & 10.05 $\pm$ 9.35 \\
    w/o NZO & 0.8534 & 18.24 & 0.1302 & 0.9214 & 0.7707 & 3.328 & 3.372 $\pm$ 0.30 & 8.05 $\pm$ 2.59 \\
    w/o SANAN & 0.8526 & 18.28 & 0.1323 & 0.9120 & 0.6997 & 8.453 & 8.283 $\pm$ 0.83 & 42.02 $\pm$ 4.10 \\
    \bottomrule
  \end{tabular}}
\end{table}

\subsection{Ablation Study}

To validate the effectiveness of each component in RealSkin, we conduct comprehensive ablation studies on our evaluation dataset, with quantitative results reported in Table~\ref{ablation-wo-table} and Fig.~\ref{fig: ablation_cf}, and qualitative comparisons presented in Fig.~\ref{fig: q_ablation}. Further analyses are provided in the supplementary material.

\paragraph{Effect of Semantic Features.} Removing DINO features leads to a significant drop in the DINO metric (0.8527 → 0.5242), as shown in Table~\ref{ablation-wo-table}. As illustrated in Fig.~\ref{fig: q_ablation}, this leads to unreliable initial correspondences, resulting in severe appearance distortion in the transferred results, indicating the importance of semantic guidance.

\paragraph{Effect of SGR} Without SGR, the Flip\% increases from 4.04 to 10.05 (Table~\ref{ablation-wo-table}). As shown in Fig.~\ref{fig: q_ablation}, this leads to spatially incoherent correspondences, demonstrating the role of SGR in enforcing geometric consistency.

\paragraph{Effect of NZO Refinement.} Disabling NZO increases mCD from 2.632 to 3.328 (Table~\ref{ablation-wo-table}). As further illustrated in Fig.~\ref{fig: ablation_cf}, the correspondence becomes less stable, showing that iterative spectral refinement improves robustness.

\paragraph{Effect of SANAN} Removing SANAN causes severe degradation across all metrics (Flip\%: 4.04 → 42.02, mCD: 2.632 → 8.453), as reported in Table~\ref{ablation-wo-table}. As shown in Fig.~\ref{fig: q_ablation}, the resulting transfer becomes distorted, confirming the necessity of neural spectral modeling.

\paragraph{Summary.}
Overall, these results demonstrate that each component contributes positively to the final performance. Semantic features improve initialization robustness, SGR enforces spatial coherence, NZO stabilizes spectral refinement, and SANAN enables expressive non-isometric modeling. The full system integrates these components into a unified framework that consistently improves both geometric accuracy and appearance fidelity.

\begin{figure}
    \centering
    \includegraphics[width=1.0\linewidth]{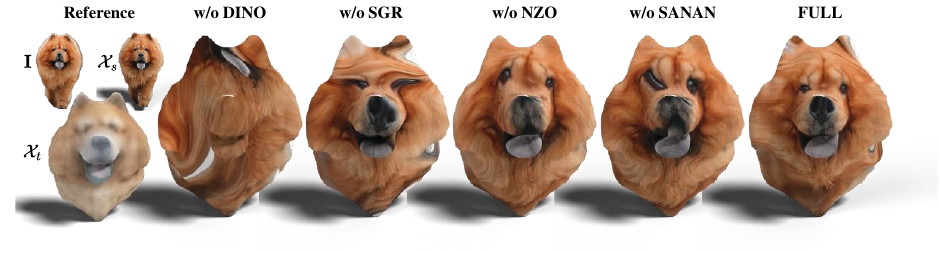}
    \caption{Qualitative ablation comparisons of attribute transfer. The full RealSkin model successfully maintains spatial continuity and achieves accurate, smooth attribute transfer.}
    \label{fig: q_ablation}
\end{figure}

\section{Conclusion and Future Work}

We have presented RealSkin, a functional map-based framework for real-to-synthetic appearance transfer that enables physically grounded attribute propagation from images to 3D assets. By integrating spatial-guided registration with neural spectral optimization, our method effectively handles topological discrepancies and non-isometric deformations, which are common in real-world scenarios. Extensive experiments demonstrate that RealSkin consistently outperforms existing shape correspondence and appearance enhancement methods in both geometric accuracy and visual fidelity.

Despite its effectiveness, RealSkin has several limitations. First, its performance is sensitive to the quality of initial geometry reconstruction and feature extraction, particularly under heavy occlusions or texture ambiguity. Second, since the current formulation focuses on partial correspondence, global texture consistency across the entire object is not explicitly enforced.

In future work, we plan to explore UV-space completion and global texture optimization strategies to achieve fully consistent appearance transfer. Additionally, extending the framework to handle dynamic scenes and non-rigid temporal sequences would further broaden its applicability in real-world 3D content generation.

\clearpage
% DO NOT INCLUDE ACKNOWLEDGMENTS IN AN ANONYMOUS SUBMISSION TO SIGGRAPH 2019
%\begin{acks}
%
%The authors would like to thank Dr. Maura Turolla of Telecom
%Italia for providing specifications about the application scenario.
%
%The work is supported by the \grantsponsor{GS501100001809}{National
%  Natural Science Foundation of
%  China}{http://dx.doi.org/10.13039/501100001809} under Grant
%No.:~\grantnum{GS501100001809}{61273304\_a}
%and~\grantnum[http://www.nnsf.cn/youngscientists]{GS501100001809}{Young
%  Scientists' Support Program}.
%
%
%\end{acks}

% Bibliography
\bibliographystyle{ACM-Reference-Format}
\bibliography{sample-bibliography}

% Appendix
\appendix

\begin{figure*}
    \centering
    \includegraphics[width=0.9\linewidth]{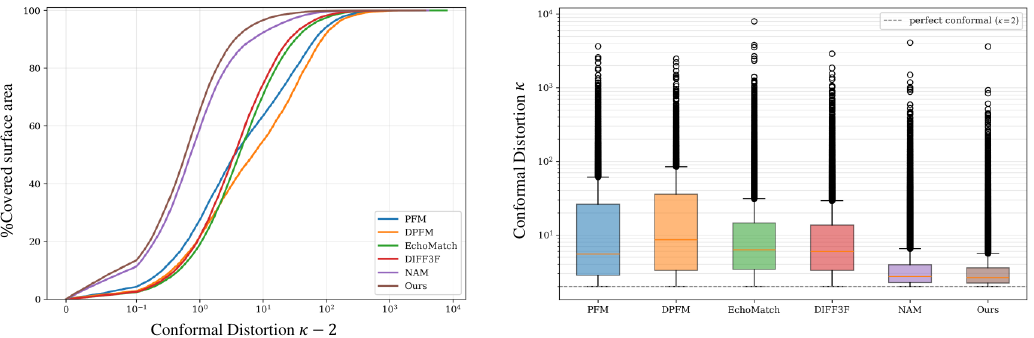}
    \caption{Quantitative comparison of conformal distortion. Left: Area-weighted cumulative distribution function (CDF) of conformal distortion ($\kappa-2$); our curve stays closest to the top-left, indicating minimal distortion across larger areas. Right: Per-triangle distribution of $\kappa$ (log scale). Compared to baselines, our method yields a median closer to the ideal $\kappa=2$ and a tighter inter-quartile range, demonstrating superior and more uniform angle preservation.}
    \label{fig: Comparison_conformal}
\end{figure*}

\begin{figure*}
    \centering
    \includegraphics[width=0.9\linewidth]{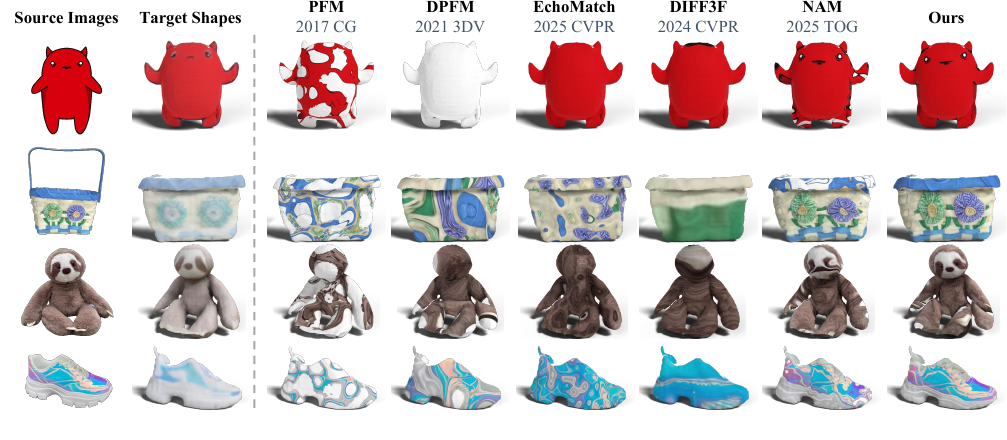}
    \caption{Qualitative comparison with functional map-based methods. Our approach demonstrates superior transfer quality, even in the presence of topological discrepancies and non-isometric deformations.}
    \label{fig: qualitative_comparison}
\end{figure*}

\begin{figure*}
    \centering
    \includegraphics[width=0.9\linewidth]{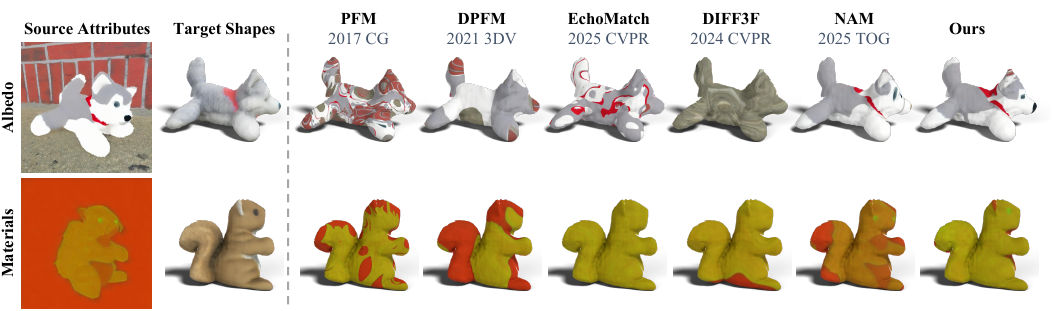}
    \caption{Qualitative results showing the transfer of different physical attributes. Our method achieves high-fidelity physical attribute transfer onto geometric surfaces.}
    \label{fig: attribute_transfer}
\end{figure*}

\begin{figure*}
    \centering
    \includegraphics[width=0.9\linewidth]{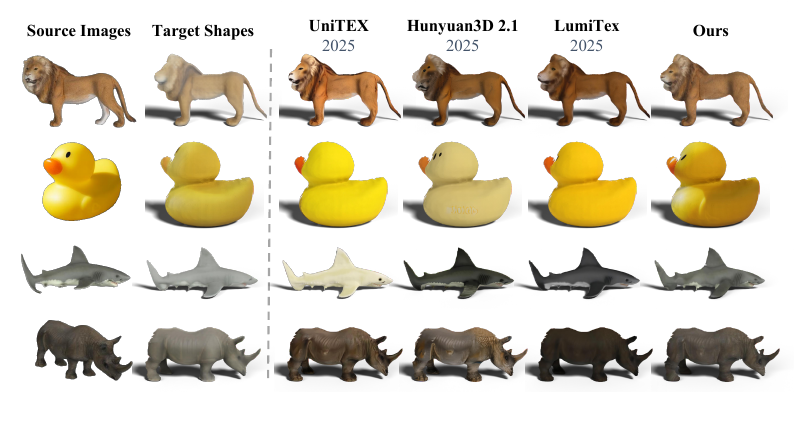}
    \caption{Qualitative comparison with generative methods. By directly transferring attributes from the source image, our approach achieves higher fidelity and better preserves fine details.}
    \label{fig: qualitative_comparison_gen}
\end{figure*}

\begin{figure*}
    \centering
    \includegraphics[width=0.8\linewidth]{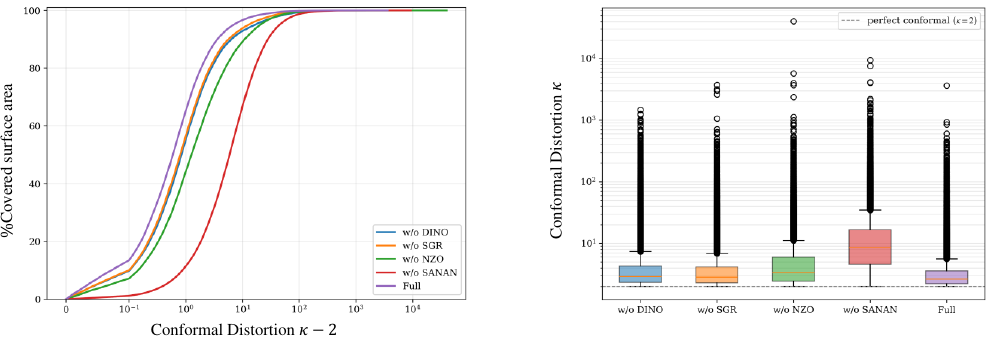}
    \caption{Quantitative ablation of mapping smoothness.}
    \label{fig: ablation_cf}
\end{figure*}

\begin{figure*}
    \centering
    \includegraphics[width=0.7\linewidth]{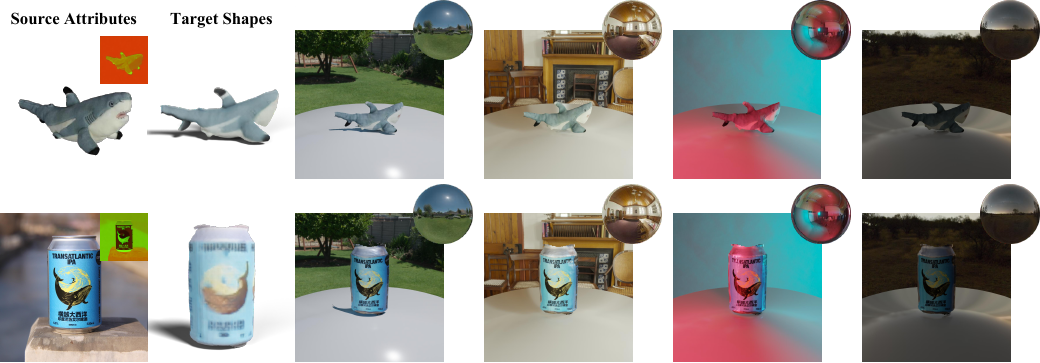}
    \caption{Visualization of joint appearance and physical attribute transfer results under different HDRI maps.}
    \label{fig: relighting}
\end{figure*}

\end{document}